\documentclass[sigconf]{acmart}

\usepackage{multirow}
\usepackage{algorithm}
\usepackage{algorithmic}
\usepackage{subfigure}

\AtBeginDocument{%
  }

\copyrightyear{2023}
\acmYear{2023}
\setcopyright{acmlicensed}\acmConference[CIKM '23]{Proceedings of the 32nd ACM International Conference on Information and Knowledge Management}{October 21--25, 2023}{Birmingham, United Kingdom}
\acmBooktitle{Proceedings of the 32nd ACM International Conference on Information and Knowledge Management (CIKM '23), October 21--25, 2023, Birmingham, United Kingdom}
\acmPrice{15.00}
\acmDOI{10.1145/3583780.3614999}
\acmISBN{979-8-4007-0124-5/23/10}

\begin{document}

\title{Optimal Linear Subspace Search: Learning to Construct Fast and High-Quality Schedulers for Diffusion Models}

\author{Zhongjie Duan}
\affiliation{
  \institution{East China Normal University}
  \city{Shanghai}
  \country{China}
}
\email{zjduan@stu.ecnu.edu.cn}

\author{Chengyu Wang}
\affiliation{
  \institution{Alibaba Group}
  \city{Hangzhou}
  \country{China}
}
\email{chengyu.wcy@alibaba-inc.com}

\author{Cen Chen}\authornote{Corresponding author.}
\affiliation{
  \institution{East China Normal University}
  \city{Shanghai}
  \country{China}
}
\email{cenchen@dase.ecnu.edu.cn}

\author{Jun Huang}
\affiliation{
  \institution{Alibaba Group}
  \city{Hangzhou}
  \country{China}
}
\email{huangjun.hj@alibaba-inc.com}

\author{Weining Qian}
\affiliation{
  \institution{East China Normal University}
  \city{Shanghai}
  \country{China}
}
\email{wnqian@dase.ecnu.edu.cn}

\renewcommand{\shortauthors}{Duan et al.}

\begin{abstract}
In recent years, diffusion models have become the most popular and powerful methods in the field of image synthesis, even rivaling human artists in artistic creativity. However, the key issue currently limiting the application of diffusion models is its extremely slow generation process. Although several methods were proposed to speed up the generation process, there still exists a trade-off between efficiency and quality. In this paper, we first provide a detailed theoretical and empirical analysis of the generation process of the diffusion models based on schedulers. We transform the designing problem of schedulers into the determination of several parameters, and further transform the accelerated generation process into an expansion process of the linear subspace. Based on these analyses, we consequently propose a novel method called Optimal Linear Subspace Search (OLSS), which accelerates the generation process by searching for the optimal approximation process of the complete generation process in the linear subspaces spanned by latent variables. OLSS is able to generate high-quality images with a very small number of steps. To demonstrate the effectiveness of our method, we conduct extensive comparative experiments on open-source diffusion models. Experimental results show that with a given number of steps, OLSS can significantly improve the quality of generated images. Using an NVIDIA A100 GPU, we make it possible to generate a high-quality image by Stable Diffusion within only one second without other optimization techniques.
\end{abstract}




\maketitle

\section{Introduction}

\begin{figure}
\begin{center}
\includegraphics[width=1.0\linewidth]{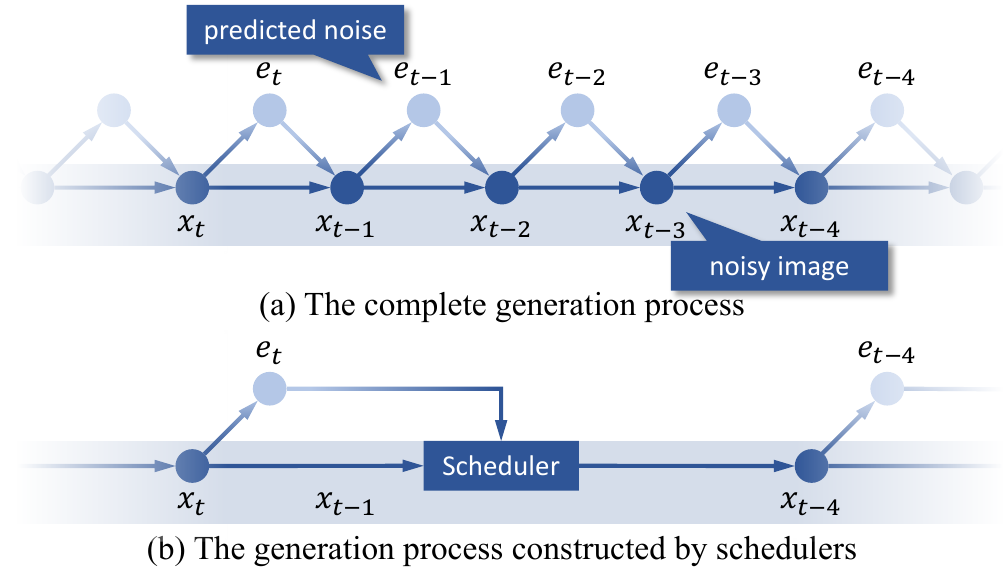}
\end{center}
\caption{The complete generation process of diffusion models consists of hundreds of steps for gradual denoising. Diffusion schedulers speed up this process by skipping some steps but may make destructive changes to images.}
\label{fig:schedulers}
\end{figure}

In recent years, diffusion models \cite{ho2020denoising, nichol2021improved, song2019generative} have become the most popular framework and have achieved impressive success in image synthesis \cite{feng2022ernie, ramesh2022hierarchical, saharia2022photorealistic}. Unlike generative adversarial networks (GANs) \cite{goodfellow2020generative}, diffusion models generate high-quality images without relying on adversarial training processes, and do not require careful hyperparameter tuning. In addition to image synthesis, diffusion models have also been applied to image super-resolution \cite{saharia2022image}, music synthesis \cite{liu2021diffsinger} and video synthesis \cite{esser2023structure}. The research community and industry have witnessed the impressive effectiveness of diffusion models in generative tasks.

The major weakness of diffusion models, however, is the extremely slow sampling procedure, which limits their practicability \cite{yang2022diffusion}. The diffusion model is a family of iterative generative models and typically requires hundreds to thousands of steps to generate content. The idea of diffusion models is inspired by the diffusion process in physics. In image synthesis, various levels of Gaussian noise are incrementally added to images (or latent variables corresponding to images) in a stepwise manner, while the model is trained to denoise the samples and reconstruct the original images. The generation process is the reverse process of diffusion process and both processes include the same time discretization steps to ensure consistency. The number of steps in the training stage required is usually very large to improve the image quality, making the generation process extremely slow.

To tackle the efficiency problem of diffusion models, existing studies proposed several types of methods. For instance, Salimans et al. \cite{salimans2022progressive} proposed a distillation method that can reduce the number of sampling steps to half and can be applied repeatably to a model. By maximizing sample quality scores, DDSS \cite{watson2022learning} attempted to optimize fast samplers. However, these speedup methods require additional training, which makes it difficult to deploy them with limited computing resources. Another family of studies focused on designing a scheduler to control the generation process, including DDIM \cite{song2020denoising}, PNDM \cite{liu2022pseudo}, DEIS \cite{zhang2022fast}, etc. As shown in Figure \ref{fig:schedulers}, these schedulers can reduce the number of steps without training. However, when we use a small number of steps, these methods may introduce new noise and make destructive changes to images because of the inconsistency between the reconstructed generation process and the complete generation process. Hence we are still faced with a trade-off between the computing resource requirement and performance.

In order to speed up the generation process, we focus on devising an adaptive scheduler that requires only a little time for learning. In this paper, we first theoretically analyze the generation process of diffusion models. With the given steps, most of the existing schedulers \cite{song2020denoising, karras2022elucidating, liu2022pseudo} generate the next sample in the linear subspace spanned by the previous samples and model outputs, thus the generation process is essentially the expansion process of linear subspaces. We also analyze the correlation of intermediate variables when generating images, and we observe that the model outputs contain redundant duplicate information at each step. Therefore, the number of steps must be reduced to prevent redundant calculation. In light of these analyses, we replace the coefficients in the iterative formula with trainable parameters to control the expansion of the linear subspaces, and then use simple least square methods \cite{bjorck1990least, kerr2009qr} to solve these parameters. Leveraging a path optimization algorithm, we further improve the performance by tuning the sampling steps. Path optimization and parameter solving can be performed within only a few minutes, thus it is easy to train and deploy. Our proposed scheduler, named Optimal Linear Subspace Search (OLSS), is designed to be lightweight and adaptive. We apply OLSS to several popular open-source diffusion models \cite{rombach2022high} and compare the performance of OLSS with the state-of-the-art schedulers. Experimental results prove that the approximate generation process built by OLSS is an accurate approximation of the complete generation process. The source code of our proposed method has been released on GitHub\footnote{\url{https://github.com/alibaba/EasyNLP/tree/master/diffusion/olss_scheduler}}. The main contribution of this paper includes:
\begin{itemize}
    \item We theoretically and empirically analyze the generation process of diffusion models and model it as an expansion process of linear subspaces. The analysis provides valuable insights for researchers to  design new diffusion schedulers.
    \item Based on our analysis, we propose a novel scheduler, OLSS, which is capable of finding the optimal path to approximate the complete generation process and generating high-quality images with a very small number of steps.
    \item Using several popular diffusion models, we benchmark OLSS against existing schedulers. The experimental results demonstrate that OLSS achieves the highest image quality with the same number of steps.
\end{itemize}


\section{Related Work}

\subsection{Image Synthesis}

Image synthesis is an important task and has been widely investigated. In the early years, GANs \cite{reed2016generative, huang2017stacked, odena2017conditional} are the most popular methods. By adversarial training a generator and a discriminator, we can obtain a network that can directly generate images. However, it is difficult to stabilize the training process \cite{salimans2016improved}. Diffusion models overcome this issue by modeling the synthesis task as a Markovian diffusion process. Theoretically, diffusion models include Denoising Diffusion Probabilistic Models \cite{sohl2015deep}, Score-Based Generative Models \cite{song2019generative}, etc. In recent years, Latent Diffusion \cite{rombach2022high}, a diffusion model architecture that denoises images in a latent space, became the most popular model architecture. Utilizing cross-attention \cite{vaswani2017attention} and classifier-free guidance \cite{ho2022classifier}, Latent Diffusion is able to generate semantically meaningful images according to given prompts. Leveraging large-scale text-image datasets \cite{gu2022wukong, schuhmann2022laion}, diffusion models with billions of parameters have achieved impressive success in text-to-image synthesis. Diffusion models are proven to outperform GANs in image quality \cite{dhariwal2021diffusion}, and are even competitive with human artists \cite{feng2022ernie, ramesh2022hierarchical, saharia2022photorealistic}. However, the slow sampling procedure becomes a critical issue, limiting the practicability of diffusion models.

\subsection{Efficient Sampling for Diffusion Models}

The time consumed of generating an image using diffusion models is in direct proportion to the number of inference steps. To speed up the generation process, existing studies focus on reducing the number of inference steps. Specifically, some schedulers are proposed for controlling this denoising process. DDIM (Denoising Diffusion Implicit Models) \cite{song2020denoising} is a straightforward scheduler that converts the stochastic process to a deterministic process and skips some steps. Some numerical ODE algorithms \cite{butcher2000numerical, karras2022elucidating} are also introduced to improve efficiency. Liu et al. \cite{liu2022pseudo} pointed out that numerical ODE methods may introduce additional noise and are therefore less efficient than DDIM with only a small number of steps. To overcome this pitfall, they modified the iterative formula and improved their effectiveness. DEIS (Diffusion Exponential Integrator Sampler) \cite{zhang2022fast}, another study based on ODE, stabilizes the approximated generation process leveraging an exponential integrator and a semilinear structure. DPM-Solver \cite{lu2022dpm} made further refinements by calculating a part of the solution analytically. Recently, an enhanced version \cite{lu2022dpmplus} of DPM-Solver adopted thresholding methods and achieved state-of-the-art performance.

\section{Methodology}


\subsection{Review of Diffusion Models}

Different from GAN-based generative models, diffusion-based models require multi-step inference. The iterative generation process significantly increases the computation time. In the training stage, the number of steps may be very large. For example, the number of steps in Stable Diffusion \cite{rombach2022high} while training is $1000$. 

In the complete generation process, starting from random Gaussian noise $\boldsymbol x_T$, we need to calculate $\boldsymbol x_{T-1},\dots,\boldsymbol x_0$ step by step, where $T$ is the total number of steps. At each step $t$, the diffusion model $\epsilon_{\theta}$ takes $\boldsymbol x_t$ as input and output $\boldsymbol e_t=\epsilon_{\theta}(\boldsymbol x_t,t)$. We obtain $\boldsymbol x_{t-1}$ via:
\begin{equation}
\begin{split}
    \boldsymbol x_{t-1}=&\sqrt{\alpha_{t-1}}
    \underbrace{\left(\frac{\boldsymbol x_t-\sqrt{1-\alpha_t}\boldsymbol e_t}{\sqrt{\alpha_t}}\right)}_{\text{predicted }\boldsymbol x_0}\\
    &+\underbrace{\sqrt{1-\alpha_{t-1}-\sigma_t^2}\boldsymbol e_t}_{\text{direction pointing to }\boldsymbol x_t}
    +\underbrace{\sigma_t\boldsymbol\epsilon_t}_{\text{random noise}},
    \label{equation:ddpm}
\end{split}
\end{equation}
where $\alpha_t,\sigma_t$ are hyper-parameters used for training. Note that $\sigma_t\boldsymbol\epsilon_t$ is the additional random noise to increase the diversity of generated results. In DDIM \cite{song2020denoising}, $\sigma_t$ is set to $0$, making this process deterministic given $\boldsymbol x_T$.

To reduce the steps,
in most existing schedulers, a few steps $t(1),\dots,t(n)$ are selected as a sub-sequence of $\{T,T-1,\dots,0\}$, and the scheduler only calls the model to calculate $\boldsymbol e_{t(i)}$ at these $n$ steps. For example, DDIM directly transfers Formula (\ref{equation:ddpm}) to an $n$-step generation process:
\begin{equation}
\begin{split}
    \boldsymbol x_{t(i+1)}=&\sqrt{\alpha_{t(i+1)}}\left(\frac{\boldsymbol x_{t(i)}-\sqrt{1-\alpha_{t(i)}}\boldsymbol e_{t(i)}}{\sqrt{\alpha_{t(i)}}}\right)\\
    &+\sqrt{1-\alpha_{t(i+1)}}\boldsymbol e_{t(i)}.
\end{split}
\label{equation:scheduling_ddim}
\end{equation}
The final tensor $\boldsymbol x_0$ obtained by DDIM is an approximate result of that in the complete generation process. Another study \cite{karras2022elucidating} focuses on modeling the generation process as an ordinary differential equation (ODE) \cite{butcher2000numerical}. Consequently, forward Euler, a general numerical ODE algorithm, can be employed to calculate the numerical solution of $\boldsymbol x_0$:
\begin{equation}
    \boldsymbol x_{t(i+1)}=\boldsymbol x_{t(i)}+\Big(t(i+1)-t(i)\Big)\frac{\mathrm{d} \boldsymbol x_{t(i)}}{\mathrm{d}t(i)},
    \label{equation:scheduling_euler}
\end{equation}
where
\begin{equation}
    \frac{\mathrm{d} \boldsymbol x_{t}}{\mathrm{d}t}=-\frac{\mathrm{d} \alpha_t}{\mathrm{d} t}\left(
    \frac{\boldsymbol x_t}{2\alpha_t}
    -\frac{\boldsymbol e_t}{2\alpha_t\sqrt{1-\alpha_t}}
    \right).
\end{equation}
PNDM \cite{liu2022pseudo} is another ODE-based scheduler. It leverages Linear Multi-Step Method and constructs a pseudo-numerical method. We simplify the iterative formula of PNDM as:
\begin{equation}
    \boldsymbol x_{t(i+1)}=\frac{\sqrt{\alpha_{t(i+1)}}}{\sqrt{\alpha_{t(i)}}}\boldsymbol x_{t(i)}-\frac{1}{\sqrt{\alpha_{t(i)}}}\alpha_{t(i)}'\boldsymbol e_{t(i)}',
    \label{equation:scheduling_pndm}
\end{equation}
where
\begin{equation}
    \boldsymbol e_{t(i)}'=\frac{1}{24}(55 \boldsymbol e_{t(i)}-59 \boldsymbol e_{t(i-1)}+37 \boldsymbol e_{t(i-2)}-9 \boldsymbol e_{t(i-3)}),
    \label{equation:lms}
\end{equation}
\begin{equation}
    \alpha_{t(i)}'=\frac{\alpha_{t(i+1)}-\alpha_{t(i)}}
    {\sqrt{(1-\alpha_{t(i+1)})\alpha_{t(i)}}+\sqrt{(1-\alpha_{t(i)})\alpha_{t(i+1)}}}.
\end{equation}

In DDIM (\ref{equation:scheduling_ddim}) and forward Euler (\ref{equation:scheduling_euler}), $\boldsymbol x_{t(i+1)}$ is a linear combination of $\{\boldsymbol x_{t(i)},\boldsymbol e_{t(i)}\}$. In PNDM (\ref{equation:scheduling_pndm}), $\boldsymbol x_{t(i+1)}$ is a linear combination of $\{\boldsymbol x_{t(i)},\boldsymbol e_{t(i)},\boldsymbol e_{t(i-1)},\boldsymbol e_{t(i-2)},\boldsymbol e_{t(i-3)}\}$. Generally, all these schedulers satisfy 
\begin{equation}
    \boldsymbol x_{t(i+1)}\in\text{span}\{\boldsymbol x_{t(i)},\boldsymbol e_{t(1)},\boldsymbol \dots,\boldsymbol e_{t(i)}\}.
\end{equation}
Recursively, we can easily prove that
\begin{equation}
    \boldsymbol x_{t(i+1)}\in\text{span}\{\boldsymbol x_{t(1)},\boldsymbol e_{t(1)},\boldsymbol \dots,\boldsymbol e_{t(i)}\}.
    \label{equation:linear_space}
\end{equation}
Therefore, the generation process is the \textit{expansion process of a linear subspace}. This linear subspace is spanned by the initial Gaussian noise and the previous model outputs. At each step, we obtain the model output and add it to the vector set. The core issue of designing a scheduler is to determine the coefficients in the iterative formula. The number of non-zero coefficients does not exceed $\frac{1}{2}n^2+\frac{3}{2}n$.

\subsection{Empirical Analysis of Generation Process}
\label{sec:empirical_analysis}

\begin{figure}
\begin{center}
\includegraphics[width=.8\linewidth]{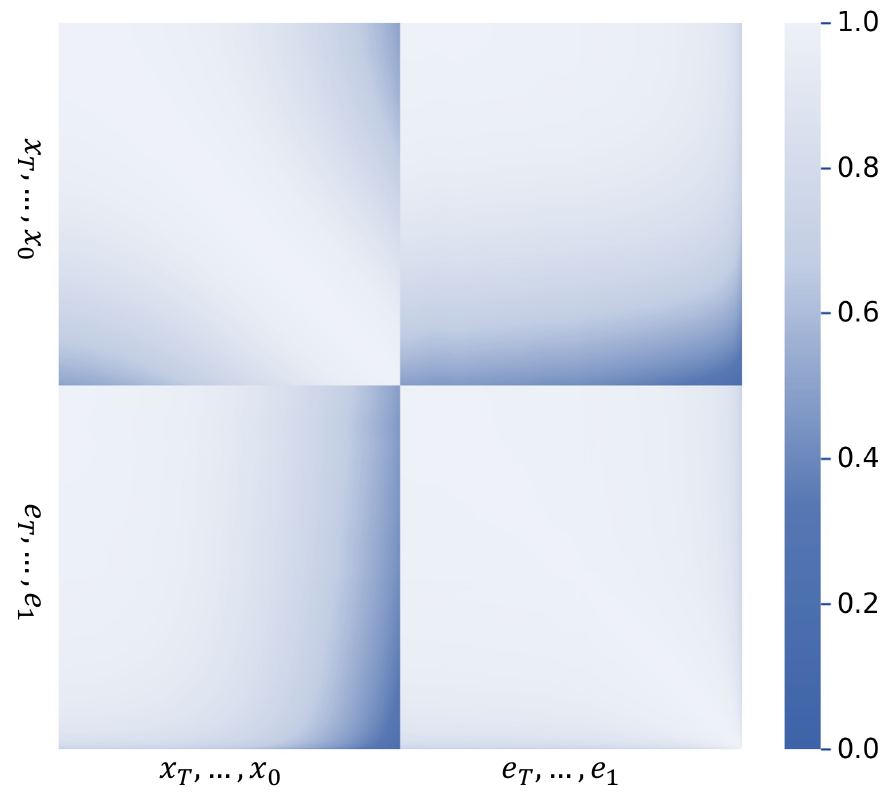}
\end{center}
\caption{The heat map of the correlation coefficients of latent variables, which records the whole generation process.}
\label{fig:heatmap}
\end{figure}

\begin{figure*}
\begin{center}
\includegraphics[width=1.0\linewidth]{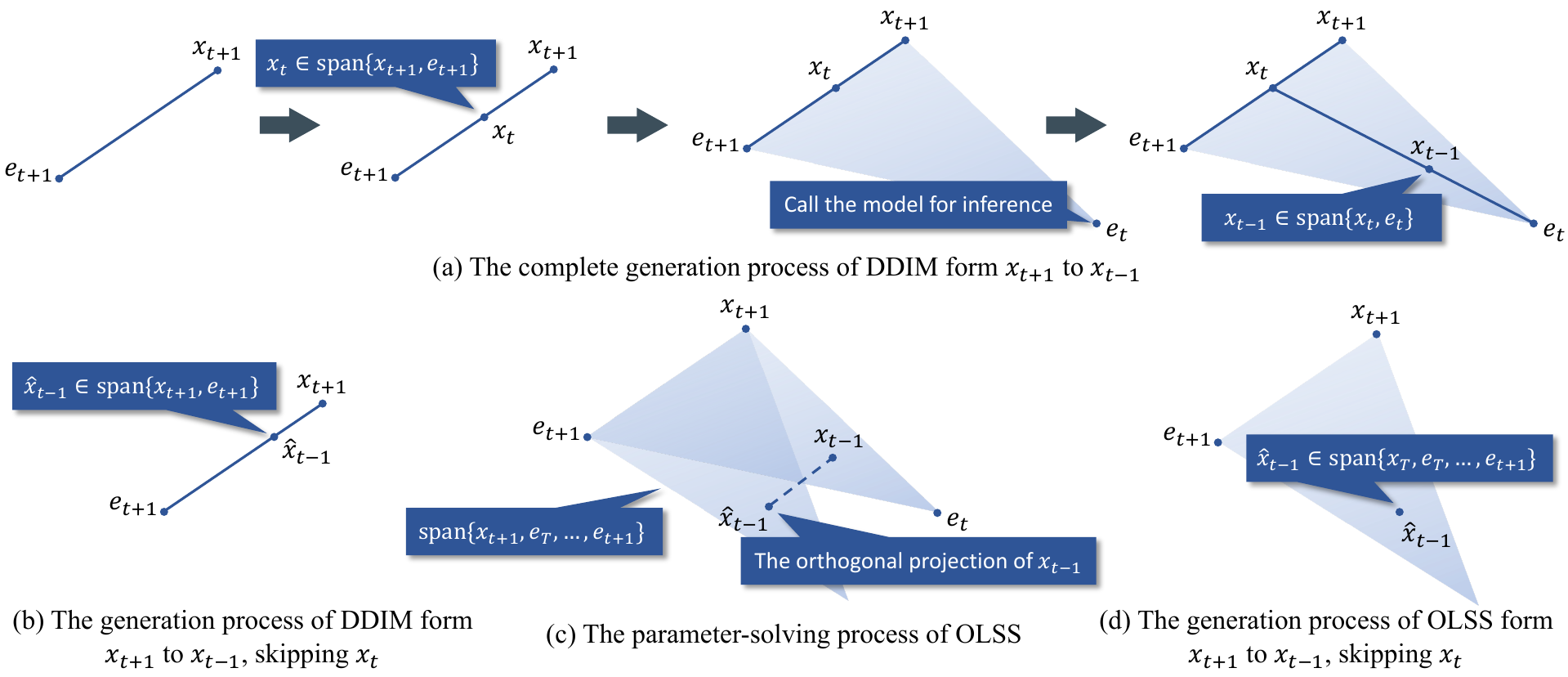}
\end{center}
\vspace{-.5em}
\caption{An interpretation of OLSS. (a) In the complete generation process of DDIM from $\boldsymbol x_{t+1}$ to $\boldsymbol x_{t-1}$, the scheduler first computes $\boldsymbol x_t\in\text{span}\{\boldsymbol x_{t+1},\boldsymbol e_{t+1}\}$ and then computes $\boldsymbol x_{t-1}\in\text{span}\{\boldsymbol x_t,\boldsymbol e_t\}$. We have $\boldsymbol x_{t-1}\in\text{span}\{\boldsymbol x_{t+1},\boldsymbol e_{t+1},\boldsymbol e_t\}$. (b) If we use DDIM to skip $\boldsymbol x_t$, the scheduler will compute $\hat{\boldsymbol x}_{t-1}\in\text{span}\{\boldsymbol x_{t+1},\boldsymbol e_{t+1}\}$ using the iterative Formula (\ref{equation:scheduling_ddim}). (c) If we use OLSS to skip $\boldsymbol x_t$, we use the orthogonal projection of $\boldsymbol x_{t-1}$ in $\text{span}\{\boldsymbol x_{t+1},\boldsymbol e_T,\dots,\boldsymbol e_{t+1}\}$ as the estimation of $\boldsymbol x_{t-1}$. In the parameter-solving process, we compute the orthogonal projection matrix and construct the new iterative formula. (d) The linear subspace spanned by $\{\boldsymbol x_{t+1},\boldsymbol e_T,\dots,\boldsymbol e_{t+1}\}$ is equivalent to the linear subspace spanned by $\{\boldsymbol x_T,\boldsymbol e_T,\dots,\boldsymbol e_{t+1}\}$. In the generation process of OLSS, we directly compute $\hat{\boldsymbol x}_{t-1}\in\text{span}\{\boldsymbol x_T,\boldsymbol e_T,\dots,\boldsymbol e_{t+1}\}$. Compared with DDIM, the estimation of $\boldsymbol x_{t-1}$ is in a higher dimensional linear subspace, thus it is more accurate.}
\label{fig:skip_connection}
\end{figure*}

To empirically analyze what happens in the whole generation process, we use Stable Diffusion to generate several images and store the latent variables, including $\boldsymbol x_T,\dots,\boldsymbol x_0,\boldsymbol e_T,\dots,\boldsymbol e_1$. As shown in Figure \ref{fig:heatmap}, we plot a heat map showing the correlation coefficients between these variables. We have the following findings:
\begin{enumerate}
\setlength{\itemsep}{0pt}
\setlength{\parsep}{0pt}
    \item $\boldsymbol x_t$ is similar to that in neighboring steps and differs from that in non-neighboring steps. As the denoising process proceeds, $\boldsymbol x_t$ is updated continuously. 
    \item The correlation between $\boldsymbol x_t$ and the predicted noise $\boldsymbol e_t$ is strong in the beginning and becomes weak in the end. The reason is that $\boldsymbol x_t$ consists of little noise in the last few steps.
    \item The correlation between $\boldsymbol e_T,\dots,\boldsymbol e_1$ is significantly strong, indicating that the outputs of the model contain redundant duplicate information.
\end{enumerate}

\subsection{Constructing a Faster Scheduler}
\label{sec:skip_connection}

On the basis of existing schedulers and our analysis, we propose OLSS, a new diffusion scheduler. In this method, we first run the complete generation process to catch the intermediate variables and then construct an approximate process using these variables, instead of leveraging mathematical theories to design a new iterative formula.

Assume that we have selected $n$ steps $\{t(1),\dots,t(n)\}$, which is a sub-sequence of $\{T,T-1,\dots,0\}$ with $t(1)=T$. Calling the diffusion model for inference is only allowed at these $n$ steps. At the $i$-th step, we have obtained intermediate variables $\boldsymbol x_{t(1)},\dots,\boldsymbol x_{t(i)}$ and $\boldsymbol e_{t(1)},\dots,\boldsymbol e_{t(i)}$. In the complete generation process, it needs to call the model for $t(i+1)-t(i)-1$ times. To reduce time consumption, we naturally come up with a naive method. As we mentioned in Section \ref{sec:empirical_analysis}, the correlation between $\boldsymbol e_T,\dots,\boldsymbol e_1$ is significantly strong, thus we can estimate the model's output $\boldsymbol e_{t(i)-1}$ using a simple linear model trained with intermediate variables. Formally, let
\begin{equation}
    \hat{\boldsymbol e}_{t(i)-1}=\mathop{\arg\min}_{\boldsymbol e\in \mathcal E}||\boldsymbol e-\boldsymbol e_{t(i)-1}||_2^2,
    \label{equation:naive_objective}
\end{equation}
where the feasible region is
\begin{equation}
    \mathcal E=\text{span}\{\boldsymbol x_{t(1)},\boldsymbol e_{t(1)},\dots,\boldsymbol e_{t(i)}\}.
    \label{equation:naive_feasible_region}
\end{equation}
In other words, we use the orthogonal projection of $\boldsymbol e_{t(i)-1}$ in $\mathcal E$ as the estimation of $\boldsymbol e_{t(i)-1}$. Similarly, we can obtain all estimated intermediate variables in the missing steps before $t(i+1)$ and finally calculate $\hat{\boldsymbol x}_{t(i+1)}$. According to Equations (\ref{equation:linear_space}, \ref{equation:naive_objective}, \ref{equation:naive_feasible_region}), it is obvious to see the estimated $\boldsymbol x_{t(i+1)}$ still satisfies Equation (\ref{equation:linear_space}). However, note that the estimation may have non-negligible errors and the errors are accumulated into subsequent steps. To address this issue, we design a simplified end-to-end method that directly estimates $\boldsymbol x_{t(i+1)}$. The simplified method only contains one linear model, containing $i+1$ coefficients $w_{i,0},w_{i,1},\dots,w_{i,i}$. The estimated $\boldsymbol x_{t(i+1)}$ is formulated as:
\begin{equation}
    \hat{\boldsymbol x}_{t(i+1)}=w_{i,0}\boldsymbol x_{t(1)}+\sum_{j=1}^i w_{i,j}\boldsymbol e_{t(j)}.
    \label{equation:skip_connection}
\end{equation}
The decision space in this simplified method is consistent with the naive method mentioned above. Leveraging least square methods \cite{bjorck1990least}, we can easily minimize the mean square error $||\hat{\boldsymbol x}_{t(i+1)}-\boldsymbol x_{t(i+1)}||_2^2$. Note that we use only $\boldsymbol x_{t(1)}$ (i.e., $\boldsymbol x_T$) instead of all intermediate variables $\{\boldsymbol x_{t(1)},\dots,\boldsymbol x_{t(i)}\}$ in Equation (\ref{equation:skip_connection}), because $\{\boldsymbol x_{t(1)},\boldsymbol e_{t(1)},\dots,\boldsymbol e_{t(i)}\}$ is a linearly independent set and the other vectors $\{\boldsymbol x_{t(2)},\dots,\boldsymbol x_{t(i)}\}$ can be linearly represented by these vectors. Additionally, the linear independence makes it easier to solve the least squares problem using QR-decomposition algorithms \cite{kerr2009qr}, which is faster and more numerically stable than directly computing the pseudo-inverse matrix \cite{regalia1993numerical}.

\begin{algorithm}[t]
    \caption{Step Searching with error upper bound $D$}
    \label{algorithm:find_step}
    \begin{algorithmic}[1]
        \STATE \textbf{Input:} Error upper bound $D$
        \STATE \textbf{Input:} Previous steps $\{t(1),\dots,t(i)\}$
        \STATE $t_l= t(i),t_r= 0$
        \WHILE{$t_l>t_r$}
            \STATE $t_m= \lfloor\frac{t_l+t_r}{2}\rfloor$
            \IF{$d(t(1),\dots,t(i),t_m)>D$}
                \STATE $t_r= t_m+1$
            \ELSE
                \STATE $t_l= t_m$
            \ENDIF
        \ENDWHILE
        \STATE $t(i+1)= t_l$
        \RETURN $t(i+1)$
    \end{algorithmic}
\end{algorithm}

\begin{algorithm}[t]
    \caption{Path searching with error upper bound $D$}
    \label{algorithm:find_path}
    \begin{algorithmic}[1]
        \STATE \textbf{Input:} Error upper bound $D$
        \STATE $t(1)= T$
        \FOR{$i=1,2,\dots,n$}
            \STATE Find $t(i+1)$ using Algorithm \ref{algorithm:find_step}
            \IF{$t(i+1)$ does not exist}
                \RETURN None
            \ENDIF
        \ENDFOR
        \IF{$t(n+1)>0$}
            \RETURN None
        \ELSE
            \RETURN $\{t(1),t(2),\dots,t(n)\}$
        \ENDIF
    \end{algorithmic}
\end{algorithm}

\begin{algorithm}[t]
    \caption{Path optimization}
    \label{algorithm:find_error_limit}
    \begin{algorithmic}[1]
        \STATE \textbf{Input:} The required absolute error $\epsilon$ of optimal error limit
        \STATE $D_l=0$
        \STATE $D_r=10$ (a sufficiently large value)
        \WHILE{$D_r-D_l>\epsilon$}
            \STATE $D_m=\frac{t_l+t_r}{2}$
            \STATE Find a path $\mathcal T$ with error limit $D_m$ using Algorithm \ref{algorithm:find_path}
            \IF{$\mathcal T$ is not None}
                \STATE $D_r=D_m$
            \ELSE
                \STATE $D_l=D_m$
            \ENDIF
        \ENDWHILE
        \STATE Find a path $\mathcal T$ with error limit $D_r$
        \RETURN $\mathcal T$
    \end{algorithmic}
\end{algorithm}

We provide an interpretation of OLSS in Figure \ref{fig:skip_connection}. When we skip $\boldsymbol x_t$, OLSS computes an estimation of $\boldsymbol x_{t-1}$ in the linear subspace spanned by the initial Gaussian noise and the previous model outputs. Essentially, this estimation is the orthogonal projection of $\boldsymbol x_{t-1}$ in the linear subspace. Each time we call the model for inference, the linear subspace is expanded by the predicted noise. Therefore, the generation process is the expansion process of the linear subspace.

\subsection{Searching for the Optimal Path}

Another problem to be addressed is how to select the $n$ steps $\{t(1),\dots,t(n)\}$. In Section \ref{sec:empirical_analysis}, we observe that the correlation between $\boldsymbol x_t$ and $\boldsymbol e_t$ is different at each step, indicating that the difficulty level of generating low-error $\boldsymbol x_t$ is also varied. In most existing schedulers, these steps are selected in $\{T,T-1,\dots,0\}$ with equal intervals. For example, in PNDM, Equation (\ref{equation:lms}) comes from the Linear Multi-Step Method, which enforces that the steps must be uniformly selected. However, there is no restriction on the selection of steps in our method. We can search for the optimal path to generate high-quality images.

For convenience, we use $d(t(1),\dots,t(i+1))$ to denote the distance from $\boldsymbol x_{t(i+1)}$ to its orthogonal projection in the linear subspace spanned by $\{\boldsymbol x_{t(1)},\boldsymbol e_{t(1)},\dots,\boldsymbol e_{t(i)}\}$ (i.e., the error of $\hat{\boldsymbol x}_{t(i+1)}$). We add an additional step $t(n+1)=0$. To find the optimal path $\mathcal T=\{t(1),\dots,t(n+1)\}$, we formulate the path optimization problem as:
\begin{equation}
    \mathcal T=\mathop{\arg\min}_{t(1),\dots,t(n+1)} \max_{i=1}^n d(t(1),\dots,t(i+1)),
\end{equation}
\begin{equation}
    \text{s.t. } T=t(1)\ge t(2)\ge \dots \ge t(n)\ge t(n+1)=0.
\end{equation}
We intend to minimize the largest error in the $n$ steps. Setting an error upper bound $D$, we hope the error of every step does not exceed $D$. Such a path always exists if $D$ is sufficiently large, thus we can use a binary search algorithm to compute the minimal error upper bound $D$ when a path exists. The pseudo-code is presented in Algorithm \ref{algorithm:find_error_limit}. In this binary search algorithm, we have to design another algorithm to check whether the path with the error upper bound $D$ exists.

According to the conclusions in Section \ref{sec:empirical_analysis}, the more steps we skip, the larger errors we have. However, if we only skip a small number of steps to reduce the error, the path will not end at $0$ within $n$ steps. Therefore, we use another binary search algorithm to search for the next step based on a greedy strategy. By skipping more steps as possible, we can find a path with the error upper bound $D$ if it exists. The pseudo-code of finding the next step with error limitation is presented in Algorithm \ref{algorithm:find_step}, and the pseudo-code of finding the path is presented in Algorithm \ref{algorithm:find_path}.

The whole path optimization algorithm includes three loops. From inner to outer, the pseudo-codes of the three loops are presented in Algorithm \ref{algorithm:find_step}-\ref{algorithm:find_error_limit}. The first one is to find the next step with an error limitation. The second one is to check if such a path exists. Algorithm \ref{algorithm:find_path} will return the path to Algorithm \ref{algorithm:find_error_limit} if it exists. The third one is to find the minimal error upper bound. Leveraging the whole path optimization algorithm, we obtain the optimal path for constructing the generation process.

\subsection{Efficiency Analysis}

We analyze the time complexity of constructing an OLSS scheduler. The most time-consuming component is solving the least square problem. We need to solve the least square problem $\mathcal O(n)$ times if the path is fixed, and $\mathcal O(n\log \frac{1}{\epsilon}\log T)$ times to perform path optimization, where $\epsilon$ is the absolute error of optimal $D$. Empirically, when we use OLSS to improve the efficiency of Stable Diffusion, usually a few minutes of computation on the CPU is sufficient to compute the optimal path and solve all the parameters $\{w_{i,j}\}$ after all the required intermediate variables are collected from the complete generation process. In the inference phase, the computation on schedulers is negligible compared to the computation on the models. The total time consumed of generating an image is in direct proportion to the number of steps.

\section{Experiments}

To demonstrate the effectiveness of OLSS, we conduct comparative experiments. We further investigate the factors that affect the quality of images generated by OLSS.

\subsection{Comparison of Diffusion Schedulers}

We compare OLSS with $7$ baseline schedulers, including its variant OLSS-P and $6$ existing schedulers. 1) \textbf{OLSS-P}: A variant of our proposed OLSS that selects the steps $\{t(1),\dots,t(n)\}$ uniformly instead of using the path optimization algorithm. 2) \textbf{DDIM} \cite{song2020denoising}: A straightforward method that directly skips some steps in the generation process. It has been widely used in many diffusion models. 3) \textbf{Euler} \cite{karras2022elucidating}: An algorithm of calculating ODEs' numerical solution. We follow the implementation of k-diffusion\footnote{https://github.com/crowsonkb/k-diffusion} and use the ancestral sampling version. 4) \textbf{PNDM} \cite{liu2022pseudo}: An pseudo numerical method that improves the performance of Linear Multi-Step method. 5) \textbf{DEIS} \cite{zhang2022fast}: A fast high-order solver designed for diffusion models. It consists of an exponential integrator and a semi-linear structure. 6) \textbf{DPM-Solver} \cite{lu2022dpm}: A high-order method that analytically computes the linear part of the latent variables. 7) \textbf{DPM-Solver++} \cite{lu2022dpmplus}: An enhanced scheduler based on DPM-Solver. It solves ODE with a data prediction model and uses thresholding methods. DPM-Solver++ is the most recent state-of-the-art scheduler.

\subsubsection{Experimental Settings}

\begin{table*}
\caption{The FID $\downarrow$ scores between diffusion schedulers with different steps. The best results are in bold, and the second best are underscored. $\dag$: the generation result of 1000-step DDIM is the most high-quality irrespective of the computational efficiency.}
\centering
\label{table:main_table}
\setlength{\tabcolsep}{5.5pt}
\renewcommand\arraystretch{1.03}
\begin{tabular}{c|cl|cccccc|c}
\hline
\multirow{2}{*}{Model}                 & \multicolumn{2}{c|}{\multirow{2}{*}{Steps /   Scheduler}}     & \multicolumn{6}{c|}{100 Steps}                                & 1000 Steps$^{\dag}$ \\ \cline{4-10} 
                                       & \multicolumn{2}{c|}{}                                         & DDIM   & Euler  & PNDM   & DEIS   & \begin{tabular}[c]{@{}c@{}}DPM\\ -Solver\end{tabular} & \begin{tabular}[c]{@{}c@{}}DPM\\ -Solver++\end{tabular} & DDIM       \\ \hline
\multirow{24}{*}{\begin{tabular}[c]{@{}c@{}}Stable Diffusion\\ ($512\times 512$)\end{tabular}}
                                       & \multicolumn{1}{c|}{\multirow{8}{*}{5 Steps}}  & DDIM         & 82.62 & 78.14 & 84.40 & 84.91 & 85.00 & 85.24 & 84.55 \\
                                       & \multicolumn{1}{c|}{}                          & Euler        & 132.19 & 111.18 & 134.62 & 135.06 & 135.28 & 135.72 & 134.28 \\
                                       & \multicolumn{1}{c|}{}                          & PNDM         & 75.53 & 89.53 & 74.29 & 75.50 & 75.31 & 75.13 & 75.41 \\
                                       & \multicolumn{1}{c|}{}                          & DEIS         & 56.31 & \textbf{55.89} & 58.44 & 58.03 & 58.21 & 58.54 & 57.68 \\
                                       & \multicolumn{1}{c|}{}                          & DPM-Solver   & 55.39 & \underline{55.94} & 57.49 & 57.09 & 57.27 & 57.59 & 56.76 \\
                                       & \multicolumn{1}{c|}{}                          & DPM-Solver++ & 54.78 & 56.11 & 56.82 & 56.41 & 56.59 & 56.91 & 56.08 \\
                                       & \multicolumn{1}{c|}{}                          & OLSS-P       & \underline{48.67} & 61.27 & \underline{48.62} & \underline{48.80} & \underline{48.77} & \underline{48.72} & \underline{48.94} \\
                                       & \multicolumn{1}{c|}{}                          & OLSS         & \textbf{48.34} & 61.11 & \textbf{48.40} & \textbf{48.61} & \textbf{48.60} & \textbf{48.57} & \textbf{48.79} \\
                                       \cline{2-10} 
                                       & \multicolumn{1}{c|}{\multirow{8}{*}{10 Steps}} & DDIM         & 52.52 & 59.87 & 53.71 & 54.37 & 54.46 & 54.66 & 54.13 \\
                                       & \multicolumn{1}{c|}{}                          & Euler        & 79.63 & 60.04 & 81.97 & 81.70 & 81.91 & 82.33 & 81.32 \\
                                       & \multicolumn{1}{c|}{}                          & PNDM         & 55.77 & 72.31 & 54.84 & 56.01 & 55.88 & 55.77 & 56.03 \\
                                       & \multicolumn{1}{c|}{}                          & DEIS         & 39.11 & \textbf{55.07} & 41.75 & 40.96 & 41.13 & 41.40 & 40.97 \\
                                       & \multicolumn{1}{c|}{}                          & DPM-Solver   & 38.93 & \underline{55.33} & 41.53 & 40.75 & 40.92 & 41.17 & 40.73 \\
                                       & \multicolumn{1}{c|}{}                          & DPM-Solver++ & 38.42 & 55.72 & 41.06 & 40.17 & 40.33 & 40.59 & 40.02 \\
                                       & \multicolumn{1}{c|}{}                          & OLSS-P       & \underline{37.33} & 59.28 & \underline{38.64} & \underline{38.36} & \underline{38.36} & \underline{38.38} & \underline{38.60} \\
                                       & \multicolumn{1}{c|}{}                          & OLSS         & \textbf{36.46} & 58.09 & \textbf{37.68} & \textbf{37.66} & \textbf{37.66} & \textbf{37.67} & \textbf{37.79} \\
                                       \cline{2-10} 
                                       & \multicolumn{1}{c|}{\multirow{8}{*}{20 Steps}} & DDIM         & 38.04 & \underline{57.14} & 39.07 & 40.90 & 40.99 & 41.17 & 41.13 \\
                                       & \multicolumn{1}{c|}{}                          & Euler        & 65.38 & \textbf{49.92} & 67.18 & 66.85 & 67.05 & 67.39 & 66.32 \\
                                       & \multicolumn{1}{c|}{}                          & PNDM         & 37.26 & 60.57 & 35.55 & 38.38 & 38.39 & 38.42 & 38.55 \\
                                       & \multicolumn{1}{c|}{}                          & DEIS         & 23.37 & 58.88 & 28.64 & 25.87 & 25.91 & 25.99 & 26.28 \\
                                       & \multicolumn{1}{c|}{}                          & DPM-Solver   & 24.38 & 60.03 & 28.73 & 26.16 & 26.12 & 26.08 & 26.72 \\
                                       & \multicolumn{1}{c|}{}                          & DPM-Solver++ & 26.21 & 61.43 & 29.61 & 27.31 & 27.20 & 27.08 & 27.94 \\
                                       & \multicolumn{1}{c|}{}                          & OLSS-P       & \underline{22.91} & 58.96 & \underline{28.20} & \underline{25.02} & \underline{25.04} & \underline{25.11} & \underline{25.42} \\
                                       & \multicolumn{1}{c|}{}                          & OLSS         & \textbf{20.78} & 58.71 & \textbf{27.26} & \textbf{22.98} & \textbf{23.04} & \textbf{23.09} & \textbf{23.23} \\
                                       \hline
\multirow{24}{*}{\begin{tabular}[c]{@{}c@{}}Stable Diffusion 2\\ ($768\times 768$)\end{tabular}}
                                       & \multicolumn{1}{c|}{\multirow{8}{*}{5 Steps}}  & DDIM         & 93.50 & 80.83 & 95.12 & 98.10 & 98.16 & 98.38 & 97.62 \\
                                       & \multicolumn{1}{c|}{}                          & Euler        & 170.79 & 119.12 & 173.09 & 175.24 & 175.28 & 175.51 & 174.43 \\
                                       & \multicolumn{1}{c|}{}                          & PNDM         & 105.08 & 99.88 & 105.82 & 108.61 & 108.54 & 108.67 & 108.28 \\
                                       & \multicolumn{1}{c|}{}                          & DEIS         & 55.88 & \textbf{58.62} & 57.30 & 59.04 & 59.20 & 59.50 & 58.65 \\
                                       & \multicolumn{1}{c|}{}                          & DPM-Solver   & 54.90 & \underline{58.77} & 56.31 & 58.02 & 58.18 & 58.48 & 57.63 \\
                                       & \multicolumn{1}{c|}{}                          & DPM-Solver++ & 54.25 & 59.13 & 55.67 & 57.30 & 57.45 & 57.75 & 56.91 \\
                                       & \multicolumn{1}{c|}{}                          & OLSS-P       & \underline{50.03} & 65.20 & \underline{51.38} & \underline{51.95} & \underline{51.87} & \underline{51.83} & \underline{51.95} \\
                                       & \multicolumn{1}{c|}{}                          & OLSS         & \textbf{49.18} & 64.89 & \textbf{50.64} & \textbf{51.17} & \textbf{51.11} & \textbf{51.07} & \textbf{51.11} \\
                                       \cline{2-10} 
                                       & \multicolumn{1}{c|}{\multirow{8}{*}{10 Steps}} & DDIM         & 55.32 & 62.05 & 56.01 & 58.74 & 58.83 & 59.03 & 58.47 \\
                                       & \multicolumn{1}{c|}{}                          & Euler        & 96.86 & 61.75 & 98.57 & 100.88 & 100.99 & 101.26 & 100.15 \\
                                       & \multicolumn{1}{c|}{}                          & PNDM         & 56.44 & 64.58 & 56.94 & 59.52 & 59.62 & 59.81 & 59.28 \\
                                       & \multicolumn{1}{c|}{}                          & DEIS         & 39.83 & \textbf{60.85} & 42.11 & 42.59 & 42.74 & 43.02 & 42.33 \\
                                       & \multicolumn{1}{c|}{}                          & DPM-Solver   & 39.70 & \underline{61.22} & 41.93 & 42.41 & 42.56 & 42.81 & 42.17 \\
                                       & \multicolumn{1}{c|}{}                          & DPM-Solver++ & 39.40 & 61.81 & 41.62 & 41.89 & 42.04 & 42.30 & 41.66 \\
                                       & \multicolumn{1}{c|}{}                          & OLSS-P       & \underline{38.39} & 64.21 & \underline{40.46} & \underline{40.70} & \underline{40.70} & \underline{40.74} & \underline{40.63} \\
                                       & \multicolumn{1}{c|}{}                          & OLSS         & \textbf{38.10} & 63.81 & \textbf{40.15} & \textbf{40.46} & \textbf{40.48} & \textbf{40.54} & \textbf{40.45} \\
                                       \cline{2-10} 
                                       & \multicolumn{1}{c|}{\multirow{8}{*}{20 Steps}} & DDIM         & 40.19 & \underline{61.44} & 40.83 & 44.12 & 44.25 & 44.42 & 44.03 \\
                                       & \multicolumn{1}{c|}{}                          & Euler        & 74.53 & \textbf{48.51} & 76.08 & 77.80 & 77.94 & 78.18 & 77.24 \\
                                       & \multicolumn{1}{c|}{}                          & PNDM         & 40.93 & 62.20 & 40.26 & 44.19 & 44.35 & 44.61 & 43.97 \\
                                       & \multicolumn{1}{c|}{}                          & DEIS         & 27.27 & 64.34 & \underline{32.74} & \underline{30.25} & \underline{30.29} & 30.39 & \underline{30.53} \\
                                       & \multicolumn{1}{c|}{}                          & DPM-Solver   & 28.18 & 65.52 & 33.47 & 30.44 & 30.40 & \underline{30.38} & 30.82 \\
                                       & \multicolumn{1}{c|}{}                          & DPM-Solver++ & 29.81 & 67.08 & 34.66 & 31.34 & 31.24 & 31.15 & 31.88 \\
                                       & \multicolumn{1}{c|}{}                          & OLSS-P       & \underline{26.74} & 63.82 & \textbf{32.17} & 30.27 & 30.31 & 30.40 & 30.59 \\
                                       & \multicolumn{1}{c|}{}                          & OLSS         & \textbf{26.44} & 63.64 & \textbf{32.17} & \textbf{29.98} & \textbf{30.01} & \textbf{30.12} & \textbf{30.27} \\
                                       \hline
\end{tabular}
\end{table*}

\begin{table}
\caption{The FID $\downarrow$ scores of diffusion schedulers with different steps on two real-world datasets. The underlying diffusion model is fine-tuned for 5,000 steps.}
\centering
\label{table:main_table_dataset}
\setlength{\tabcolsep}{4.9pt}
\renewcommand\arraystretch{1.00}
\begin{tabular}{c|l|cc}
\hline
\multirow{2}{*}{Dataset}     & \multicolumn{1}{c|}{\multirow{2}{*}{Scheduler}} & \multicolumn{2}{c}{Steps} \\ \cline{3-4} 
                             & \multicolumn{1}{c|}{}                           & 5 Steps     & 10 Steps    \\ \hline
\multirow{8}{*}{CelebA-HQ}   & DDIM                                            & 80.55       & 31.72       \\
                             & Euler                                           & 83.80       & 35.85       \\
                             & PNDM                                            & 57.63       & 25.33       \\
                             & DEIS                                            & 24.12       & 11.68       \\
                             & DPM-Solver                                      & 23.03       & 11.68       \\
                             & DPM-Solver++                                    & 21.82       & 11.44       \\
                             & OLSS-P                                          & \underline{14.24}       & \underline{11.40}       \\
                             & OLSS                                            & \textbf{11.65}       & \textbf{11.37}       \\ \hline
\multirow{8}{*}{LSUN-Church} & DDIM                                            & 109.57      & 38.55       \\
                             & Euler                                           & 216.07      & 84.21      \\
                             & PNDM                                            & 29.55       & 14.58       \\
                             & DEIS                                            & 55.27       & 13.67       \\
                             & DPM-Solver                                      & 51.66       & 12.99       \\
                             & DPM-Solver++                                    & 48.93       & 11.77       \\
                             & OLSS-P                                          & \underline{19.46}       & \underline{11.10}       \\
                             & OLSS                                            & \textbf{15.18}       & \textbf{10.21}       \\ \hline
\end{tabular}
\end{table}

The comparative experiments consist of two parts. The first part is to benchmark the speed-up effect of these schedulers on open-domain image synthesis and analyze the relationship between different schedulers. We compare the schedulers on two popular large-scale diffusion models in the research community, including Stable Diffusion\footnote{https://huggingface.co/CompVis/stable-diffusion-v1-4} and Stable Diffusion 2\footnote{https://huggingface.co/stabilityai/stable-diffusion-2-1}. The architecture of both models consists of a CLIP-based text encoder \cite{radford2021learning}, a U-Net \cite{ronneberger2015u}, and a VAE \cite{kingma2013auto}, where the U-Net is trained to capture the pattern of noise. The implementation of baseline schedulers is mainly based on Diffusers \cite{von-platen-etal-2022-diffusers}. We randomly sample $1000$ prompts in LAION-Aesthetics V2\footnote{https://laion.ai/blog/laion-aesthetics} as the conditional information input to models. The guidance scale is set to $7.0$. The second part is to further investigate the effect of these schedulers on close-domain image synthesis. We fine-tune Stable Diffusion on CelebA-HQ \cite{karras2017progressive} ($256\times 256$) and LSUN-Church \cite{yu2015lsun} ($256\times 256$) for 5000 steps respectively. CelebA-HQ is a high-quality version of CelebA, which is a human face dataset. LSUN-Church is a part of LSUN and includes photos of churches. The training and generating process is performed without textual conditional information. We generate images using the fine-tuned model and compare them with real-world images in each dataset. In both two parts of the experiments, in order to avoid the influence of random noise on the experimental results, we use the same random seed and the same pre-generated $\boldsymbol x_T$ for every scheduler. For DPM-Solver and DPM-Solver++, we use their multi-step version to bring out their best generative effects. For OLSS, we run the complete generation process to generate $32$ images and then let our algorithm construct the approximate process.

\subsubsection{Evaluation Metrics}

We compare the quality of generated images by these methods with the same number of steps. Note that the time consumed by each scheduler is different even if the number of steps is the same, but we cannot measure it accurately because it is negligible compared to the time consumed on the model. We use FID (Frechet Inception Distance) \cite{heusel2017gans}, i.e.,~the Frechet Distance of features extracted by Inception V3 \cite{szegedy2016rethinking}, to measure the similarity between two sets of images. A smaller FID indicates that the distributions of the two sets of images are more similar. In the first part, For each scheduler, we run the generation program with 5, 10, and 20 steps respectively. Considering that the generation process with fewer steps is an approximate process of the complete generation process, we compare each one with the complete generation process (i.e., DDIM, 1000 steps). Additionally, we also compare each one with the $100$-step schedulers to investigate the consistency. In the second part, we compute the FID scores between 10,000 generated images and real-world images. The experimental results of the two parts are shown in Table \ref{table:main_table} and Table \ref{table:main_table_dataset}.

\subsubsection{Experimental Results}

In Table \ref{table:main_table}, we can clearly see that OLSS reaches the best performance with the same steps. The FID between images generated by OLSS and those by 1000-step DDIM is lower than other schedulers, and the gap is significant when we use only 5 steps. Considering the consistency of different schedulers, we observe that most schedulers generate similar images with the same settings except Euler. The FID of Euler is even larger than that of DDIM. This is because Euler's method computes the numerical solution of $\boldsymbol x_0$ iteratively along a straight line \cite{liu2022pseudo}, making the solution far from the origin $\boldsymbol x_0$. PNDM, another ODE-based scheduler, overcomes this pitfall by adjusting the linear transfer part in the iterative formula. Comprehensively, DPM-Solver++ performs the best among all baseline methods but still cannot outperform our method. Comparing OLSS and OLSS-P, we find that OLSS performs better than OLSS-P. It indicates that the path optimization algorithm further improves the performance.

In Table \ref{table:main_table_dataset}, the FID scores of OLSS are also the lowest. Even without the path optimization algorithm, OLSS-P can still outperform other schedulers. The gap between OLSS and other schedulers is more significant than that in Table \ref{table:main_table}, indicating that OLSS is more effective in generating images with a similar style. The main reason is that the generation process of images in a close domain follows a domain-specific pattern, and the learnable parameters in OLSS make it more suitable for the generation task. Additionally, the FID of PNDM is lower than DDIM, which is different from the first part of the experiments. We suspect that PNDM constructs a new generation pattern to generate realistic images rather than constructing an approximate process of the original process.

\subsection{Efficiency Study}

\begin{figure}
\begin{center}
\includegraphics[width=1.0\linewidth, trim=0 5 0 5,clip]{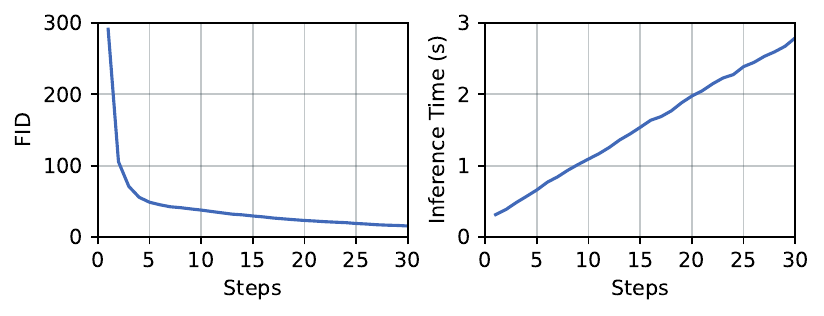}
\end{center}
\caption{The FID scores (between images generated by our method and the complete generation process) and inference time of OLSS.}
\label{fig:efficiency}
\end{figure}

When a scheduler is applied to a diffusion model, there is a trade-off between efficiency and quality. Fewer steps can reduce the inference time but usually results in lower image quality. With the same settings as above, we apply OLSS to Stable Diffusion and calculate the FID scores with varying numbers of steps. We run the program on an NVIDIA A100 GPU and record the average time of generating an image. The results are plotted in Figure \ref{fig:efficiency}. As the number of steps increases, the FID score decreases rapidly at the beginning and gradually converges. The inference time increases almost linearly with the number of steps. Setting the number of steps to $5$, OLSS is able to generate an image within only \textit{one second}, while still achieving satisfactory quality.

\subsection{Visualization}

\begin{figure}
\begin{center}
\includegraphics[width=0.8\linewidth, trim=0 5 0 5,clip]{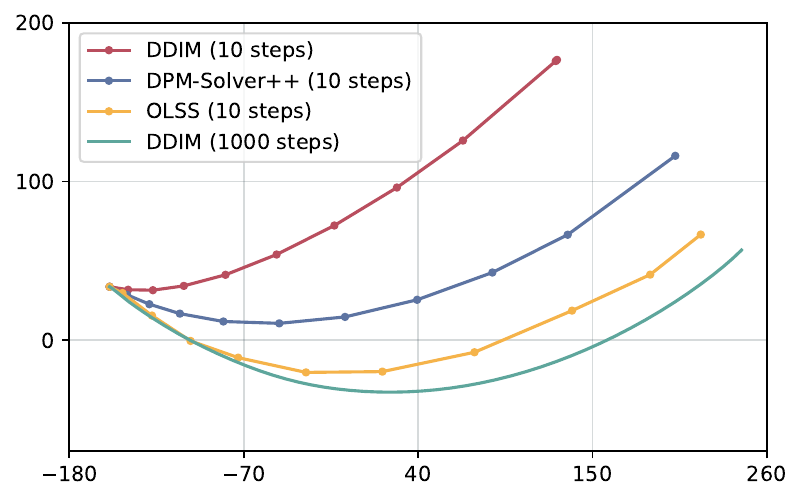}
\end{center}
\caption{The generation path $\{\boldsymbol x_{t(0)},\boldsymbol x_{t(1)},\dots,\boldsymbol x_{t(n)}\}$ of three schedulers. We embed the latent variables to 2D using PCA to see the generation process intuitively.
}
\label{fig:visualization}
\vspace{-1em}
\end{figure}

To intuitively see how diffusion models generate images with different schedulers, we select $32$ examples randomly generated by Stable Diffusion in the above experiments. We catch the generation path $\{\boldsymbol x_{t(0)},\boldsymbol x_{t(1)},\dots,\boldsymbol x_{t(n)}\}$ and then embed these latent variables into a 2D plane using Principal Component Analysis (PCA). The embedded generation paths of three schedulers are shown in Figure \ref{fig:visualization}. Starting from the same Gaussian noise $\boldsymbol x_{t(0)}$, these generation processes finally reach different $\boldsymbol x_{t(n)}$. In the complete generation process, $\boldsymbol x_i$ is updated gradually along a curve. The three 10-step schedulers construct an approximate generation process. We can see that the errors in the beginning steps are accumulated in the subsequent steps, thus the errors at the final steps become larger than those at the beginning. The generation path of OLSS is the closest one to the complete generation process, and the generation path of DPM-Solver++ is the second closest.

\begin{figure*}
\centering
\renewcommand\arraystretch{1.1}
\begin{tabular}{cccc}
\includegraphics[width=0.23\linewidth]{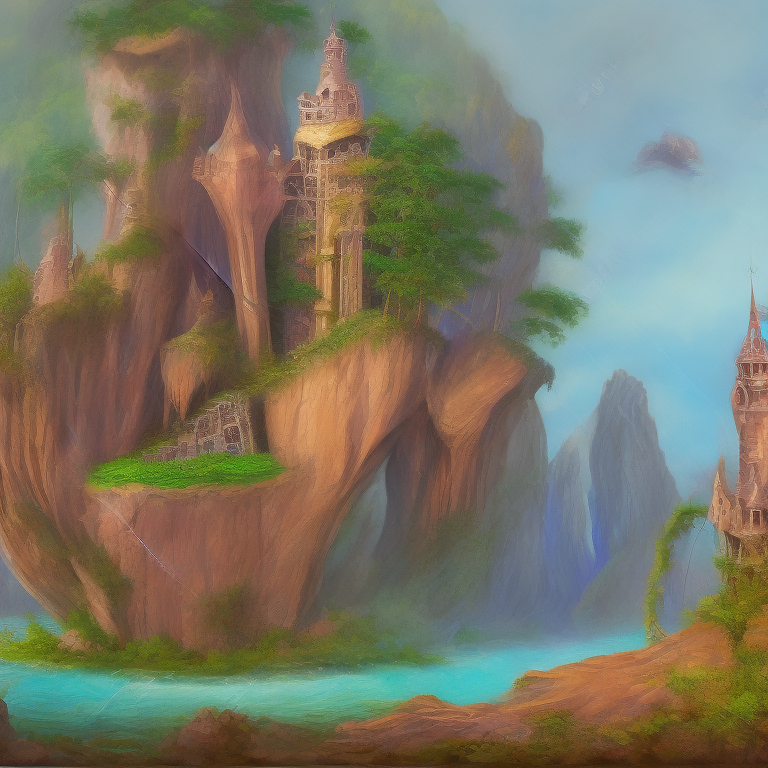} &
\includegraphics[width=0.23\linewidth]{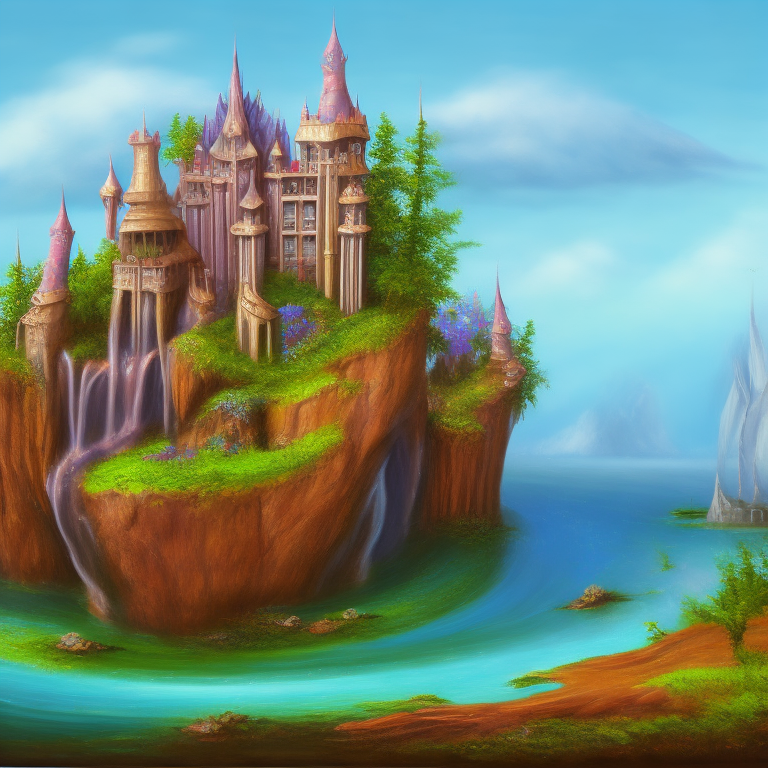} &
\includegraphics[width=0.23\linewidth]{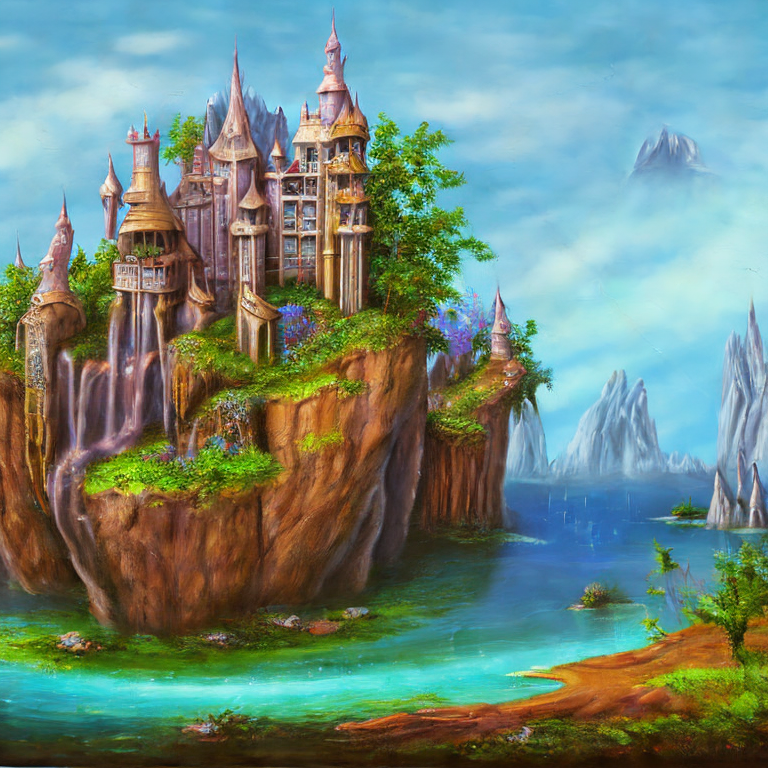} &
\includegraphics[width=0.23\linewidth]{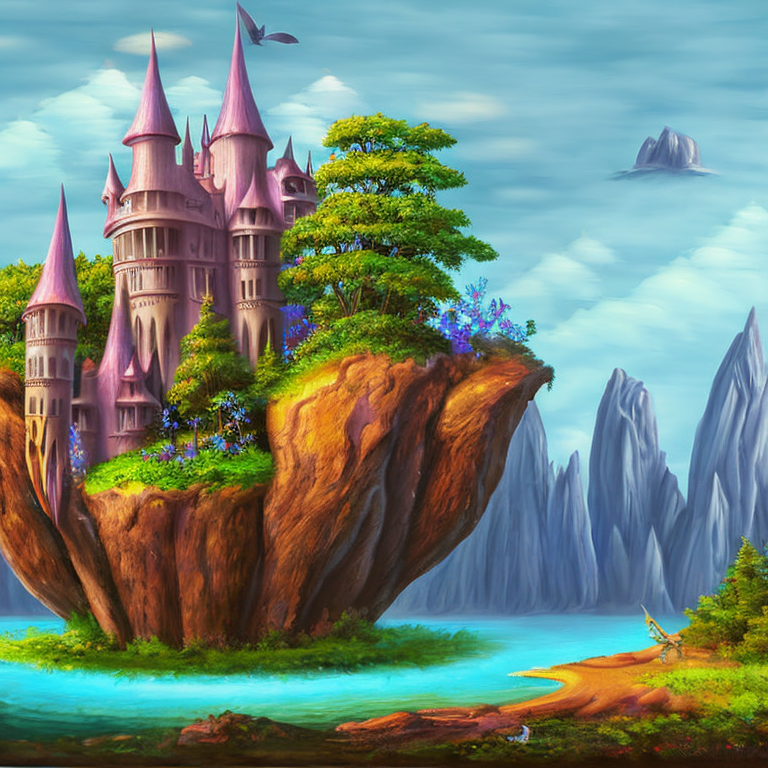}
\\
DDIM, 5 steps & DPM-Solver++, 5 steps & OLSS, 5 steps & DDIM, 1000 steps \\
\multicolumn{4}{c}{(a) Prompt: ``Fantasy magic castle on floating island. A very beautiful art painting."}
\\
\includegraphics[width=0.23\linewidth]{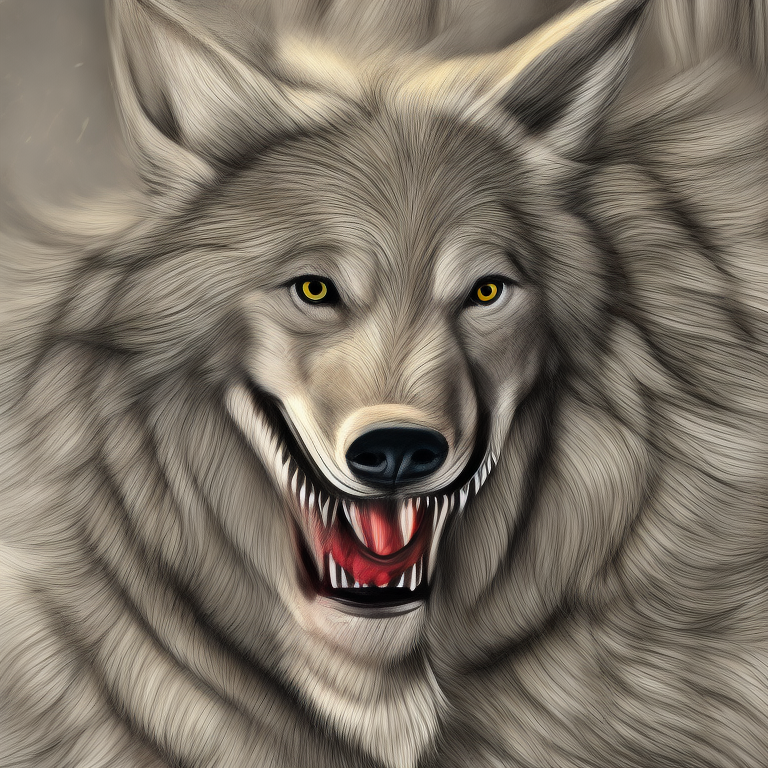} &
\includegraphics[width=0.23\linewidth]{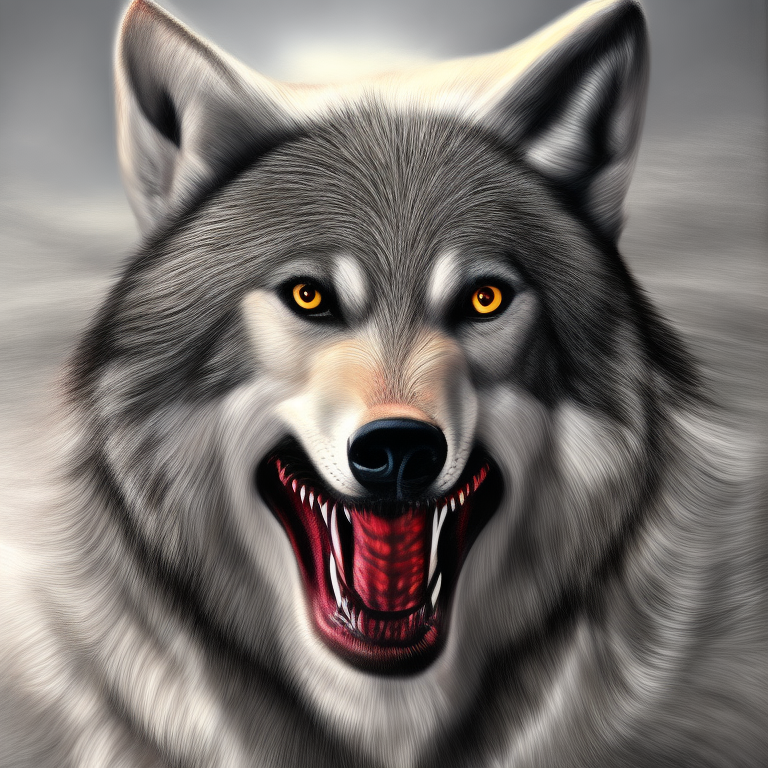} &
\includegraphics[width=0.23\linewidth]{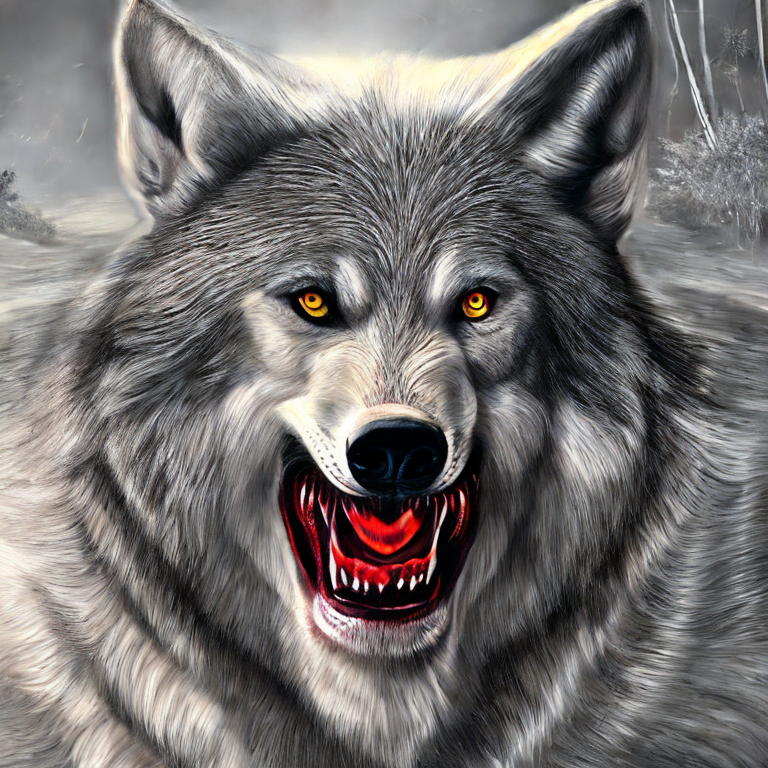} &
\includegraphics[width=0.23\linewidth]{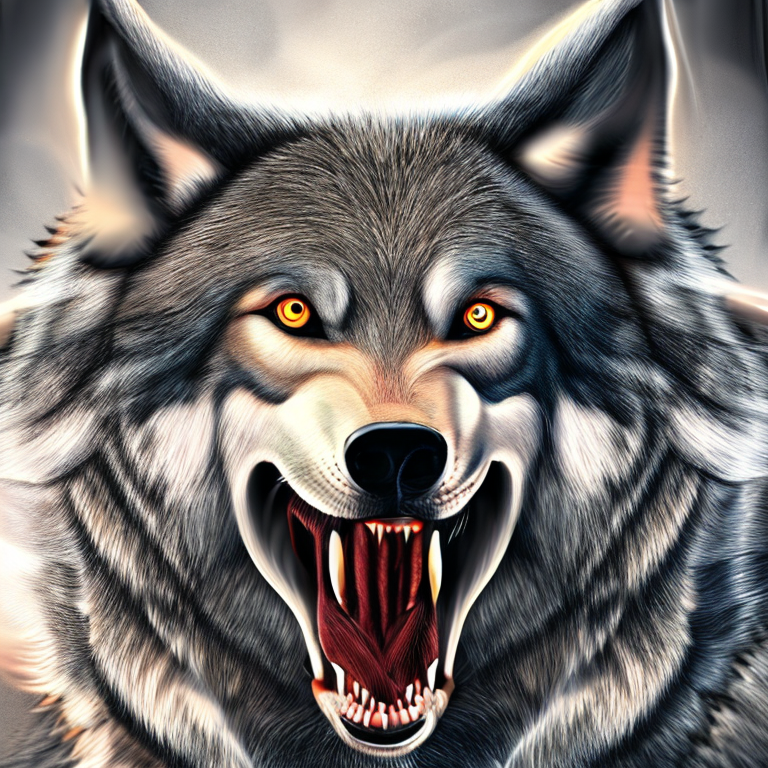}
\\
DDIM, 5 steps & DPM-Solver++, 5 steps & OLSS, 5 steps & DDIM, 1000 steps \\
\multicolumn{4}{c}{(b) Prompt: ``The leader of the wolves bared its ferocious fangs. High-resolution digital painting."}
\\
\includegraphics[width=0.23\linewidth]{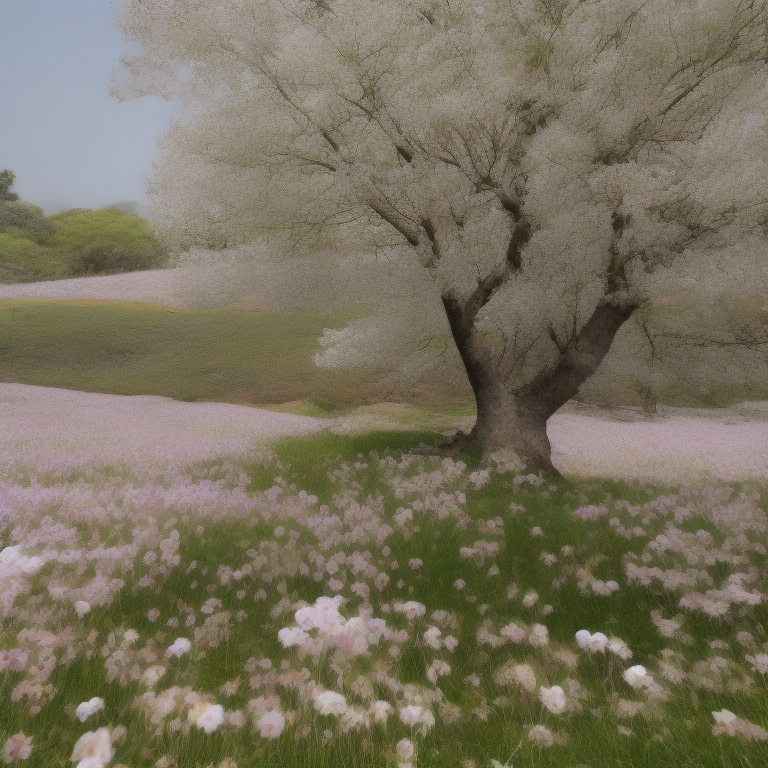} &
\includegraphics[width=0.23\linewidth]{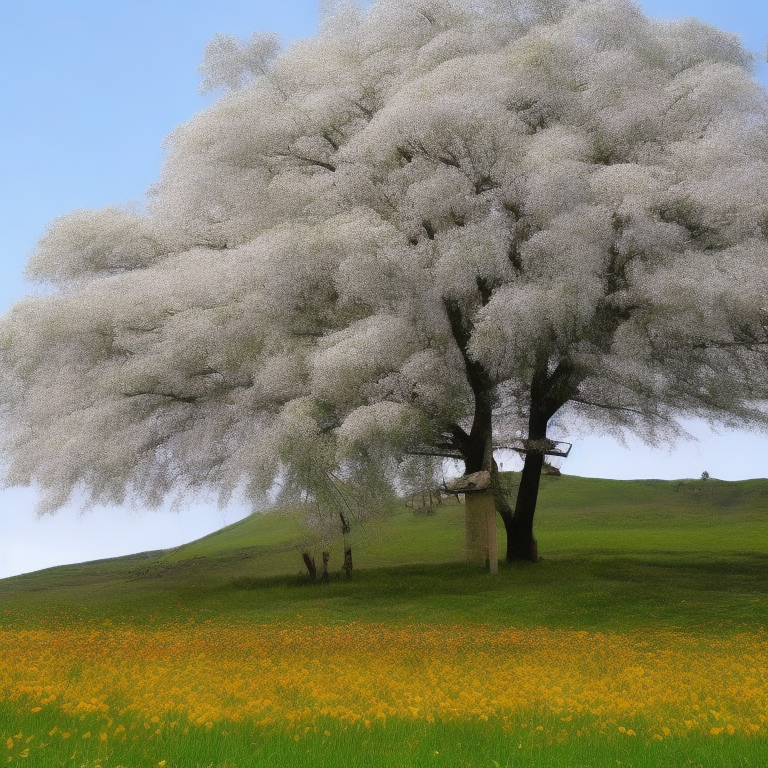} &
\includegraphics[width=0.23\linewidth]{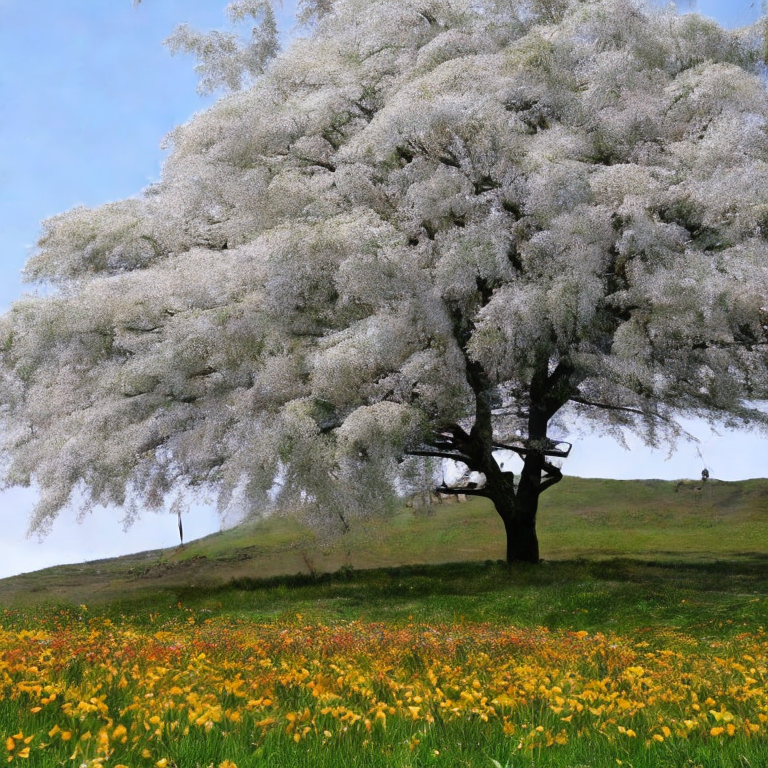} &
\includegraphics[width=0.23\linewidth]{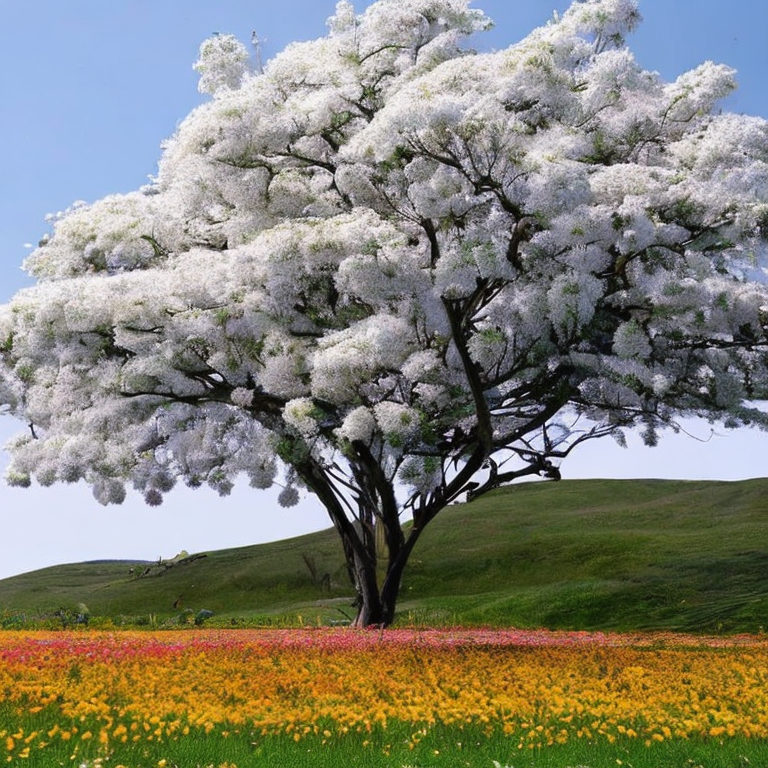}
\\
DDIM, 5 steps & DPM-Solver++, 5 steps & OLSS, 5 steps & DDIM, 1000 steps \\
\multicolumn{4}{c}{(c) Prompt: ``On the grassland, there is a towering tree with white flowers in full bloom, and under the tree are colorful flowers."}
\\
\includegraphics[width=0.23\linewidth]{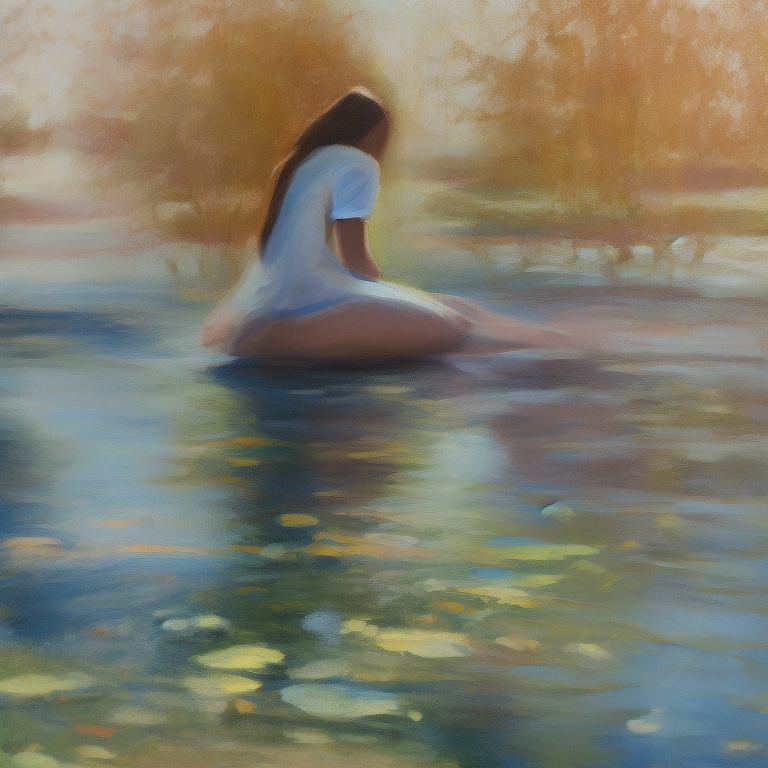} &
\includegraphics[width=0.23\linewidth]{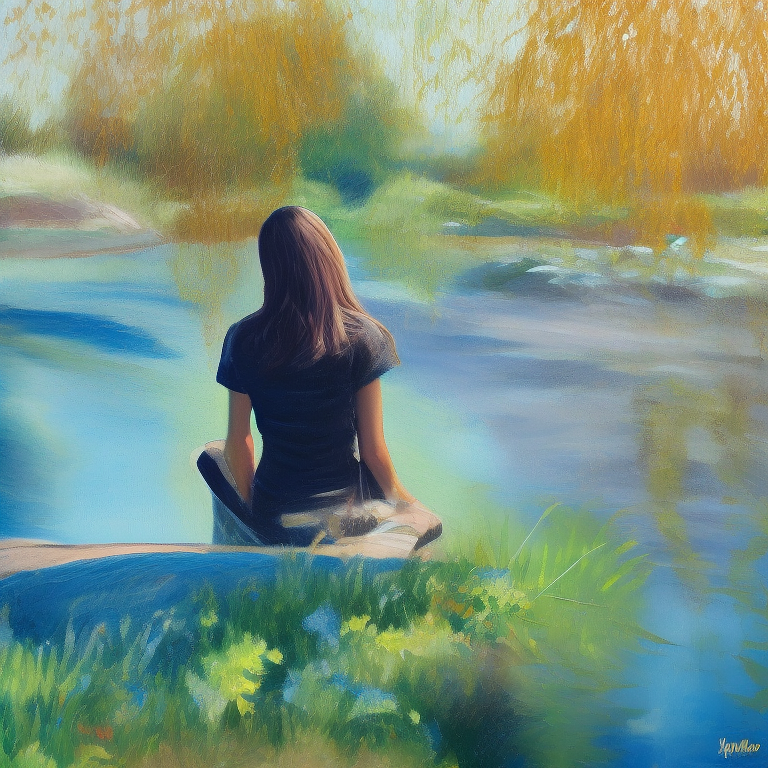} &
\includegraphics[width=0.23\linewidth]{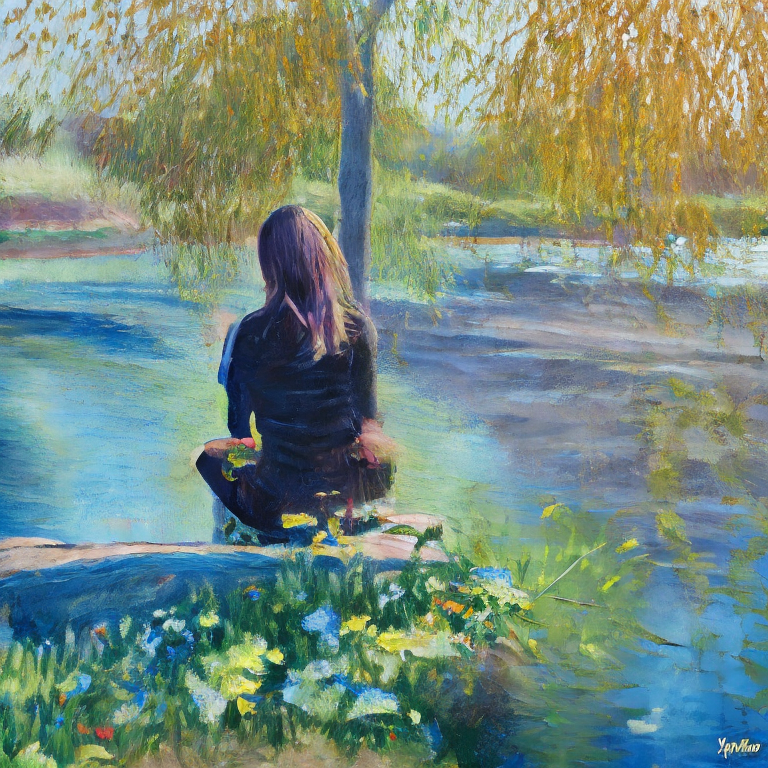} &
\includegraphics[width=0.23\linewidth]{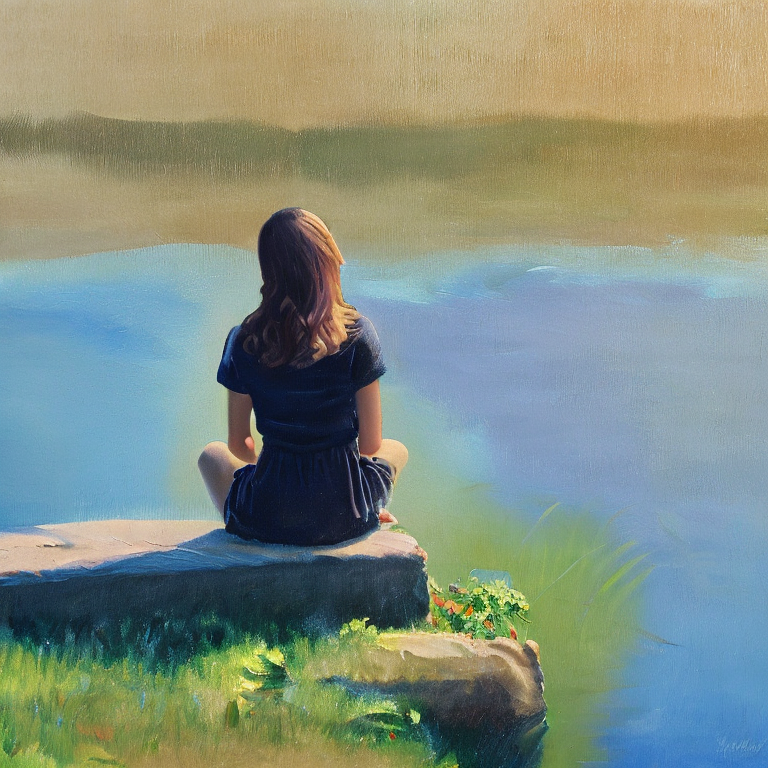}
\\
DDIM, 5 steps & DPM-Solver++, 5 steps & OLSS, 5 steps & DDIM, 1000 steps \\
\multicolumn{4}{c}{(d) Prompt: ``The girl sitting by the river looks at the other side of the river and thinks about life. Oil painting."}
\\
\end{tabular}
\caption{Some examples generated by Stable Diffusion 2 with different schedulers and steps.}
\label{fig:case_study}
\end{figure*}

\subsection{Case Study}

In Figure \ref{fig:case_study}, we present examples of images generated using Stable Diffusion 2 with DDIM, DPM-Solver++, and OLSS. Benefiting from the excellent generation ability of Stable Diffusion 2, we can generate exquisite artistic images. With only 5 steps, DDIM may make destructive changes to images. For instance, the castle in the first example looks smoggy. DPM-Solver++ tends to sharpen the entity at the expense of fuzzing details up. Only OLSS can clearly preserve the texture in the generated images (see the fur of the wolf in the second example and the flowers in the third example). In the fourth example, we observe that sometimes both DPM-Solver++ and OLSS generate images in a different style, where OLSS tends to generate more detail. Despite being generated with significantly fewer steps than 1000-step DDIM, the images generated by OLSS still look satisfactory. Hence, OLSS greatly improves the quality of generated images within only a few steps.

\section{Conclusion and Future Work}

In this paper, we investigate the schedulers in diffusion models. Specifically, we propose a new scheduler (OLSS) that is able to generate high-quality images within a very small number of steps. OLSS is a simple yet effective method that utilizes linear models to determine the coefficients in the iterative formula, instead of using mathematical theories. Leveraging the path optimization algorithm, OLSS can construct a faster process as an approximation of the complete generation process. The experimental results demonstrate that the quality of images generated by OLSS is higher than the existing schedulers with the same number of steps. In future work, we will continue investigating the generation process and explore improving the generative quality based on the modification in the latent space.

\section*{Acknowledgment}

This work was supported by the National Natural Science Foundation of China under grant number 62202170, Fundamental Research Funds for the Central Universities under grant number YBNLTS2023-014, and Alibaba Group through the Alibaba Innovation Research Program.

\bibliographystyle{ACM-Reference-Format}
\bibliography{sample-base}

\end{document}